\documentclass[graybox]{svmult}

\usepackage{helvet}         
\usepackage{courier}        
\usepackage{type1cm}        
\usepackage{mathptmx}       

\usepackage{makeidx}         
\usepackage{graphicx}        
\usepackage{multicol}        
\usepackage[bottom]{footmisc}

\usepackage{amsmath}
\usepackage{booktabs}
\usepackage{array}
\usepackage{url}
\usepackage{subfigure}
\usepackage{xcolor}
\usepackage{xspace}
\usepackage{hyperref}
\usepackage{siunitx}
\makeindex             

\begin{document}
	
	\title*{AirSim: High-Fidelity Visual and Physical Simulation for Autonomous Vehicles}
	\authorrunning{Shah, Dey, Lovett, and Kapoor}
	\author{Shital Shah$^{1}$, Debadeepta Dey$^{2}$, Chris Lovett$^{3}$, Ashish Kapoor$^{4}$}
	\institute{${1}$, ${2}$, ${3}$, ${4}$, Microsoft Research, Redmond, WA, USA \email{shitals, dedey, clovett, akapoor@microsoft.com}}
	%
	%
	\maketitle
	
	\abstract{Developing and testing algorithms for autonomous vehicles in real world is an expensive and time consuming process. Also, in order to utilize recent advances in machine intelligence and deep learning we need to collect a large amount of annotated training data in a variety of conditions and environments. We present a new simulator built on Unreal Engine that offers physically and visually realistic simulations for both of these goals. Our simulator includes a physics engine that can operate at a high frequency for real-time hardware-in-the-loop (HITL) simulations with support for popular protocols (e.g. MavLink). The simulator is designed from the ground up to be extensible to accommodate new types of vehicles, hardware platforms and software protocols. In addition, the modular design enables various components to be easily usable independently in other projects. We demonstrate the simulator by first implementing a quadrotor as an autonomous vehicle and then experimentally comparing the software components with real-world flights.}
	
	\section{Introduction}
	
	\begin{figure}[htbp]
		\centering
		\includegraphics[width=\textwidth]{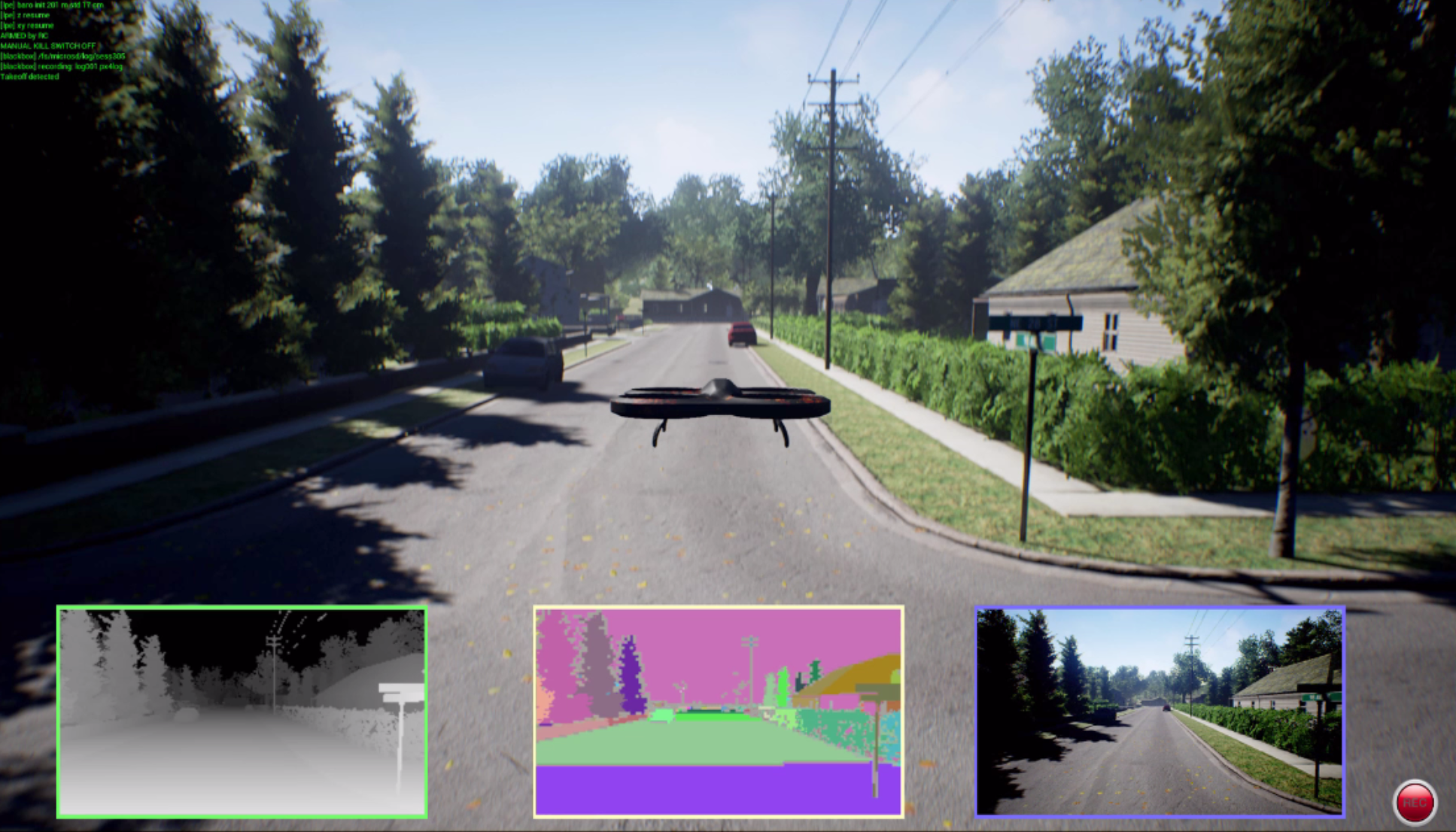}
		\caption{A snapshot from AirSim shows an aerial vehicle flying in an urban environment. The inset shows depth, object segmentation and front camera streams generated in real time.}
		\label{drone_depth_materials}
	\end{figure}
	Recently, paradigms such as reinforcement learning \cite{Kober–2013–7753}, learning-by-demonstration \cite{bagnell2015invitation} and transfer learning \cite{Weiss2016} are proving a natural means to train various robotics systems. One of the key challenges with these techniques is the high sample complexity - the amount of training data needed to learn useful behaviors is often prohibitively high. This issue is further exacerbated by the fact that autonomous vehicles are often unsafe and expensive to operate during the training phase. In order to seamlessly operate in the real world the robot needs to transfer the learning it does in simulation. Currently, this is a non-trivial task as simulated perception, environments and actuators are often simplistic and lack the richness or diversity of the real world. For example, for robots that aim to use computer vision in outdoor environments, it may be important to model  real-world complex objects such as trees, roads, lakes, electric poles and houses along with rendering that includes finer details such as soft shadows, specular reflections, diffused inter-reflections and so on. Similarly, it is important to develop more accurate models of system dynamics so that simulated behavior closely mimics the real-world.
	
	AirSim is an open-source platform \cite{AirSimGitHub} that aims to narrow the gap between simulation and reality in order to aid development of autonomous vehicles. The platform seeks to positively influence development and testing of data-driven machine intelligence techniques such as  reinforcement learning and deep learning. It is inspired by several previous simulators (see related work), and one of our key goals is to build a community to push the state-of-the-art towards this goal. 
	
	\section{Related Work}
	While an exhaustive review of currently used simulators is beyond the scope of this paper, we mention a few notable recent works that are closest to our setting and has deeply influenced this work.
	
	Gazebo \cite{koenig2004design} has been one the most popular simulation platforms for the research work. It has a modular design that allows to use different physics engines, sensor models and create 3D worlds. Gazebo goes beyond monolithic rigid body vehicles and can be used to simulate more general robots with links-and-joints architecture such as complex manipulator arms or biped robots. While Gazebo is fairly feature rich it has been difficult to create large scale complex visually rich environments that are closer to the real world and it has lagged behind various advancements in rendering techniques made by platforms such as Unreal engine or Unity.
	
	Other notable efforts includes Hector \cite{meyer2012comprehensive} that primarily focuses on tight integration with popular middleware ROS and Gazebo. It offers wind tunnel tuned flight dynamics, sensor models that includes bias drift using Gaussian Markov process and software-in-loop using Orocos toolchain. However, Hector lacks support for popular hardware platforms such as Pixhawk and protocols such as MavLink. Because of its tight dependency on ROS and Gazebo, it's limited by richness of simulated environments as noted previously.
	
	Similarly, RotorS \cite{furrer2016rotors} provides a modular framework to design Micro Aerial Vehicles, and build algorithms for control and state estimation that can be tested in simulator. It is possible to setup RotorS for HITL with Pixhawk. RotorS also uses Gazebo as its platform, consequently limiting its perception related capabilities. 
	
	Finally, jMavSim \cite{jmavsim} is easy to use simulator that was designed with a goal of testing PX4 firmware and devices. It is therefore tightly coupled with PX4 simulation APIs, uses albeit simpler sensor models and utilizes simple rendering engine without any objects in the environment. 
	
	Apart from these, there have been many games like simulators and training applications, however, these are mostly commercial closed-source software with little or no public information on models, accuracy of simulation or development APIs for autonomous applications.
	
	\section{Architecture}
\begin{figure}[ttbp]
		\includegraphics[width=\textwidth]{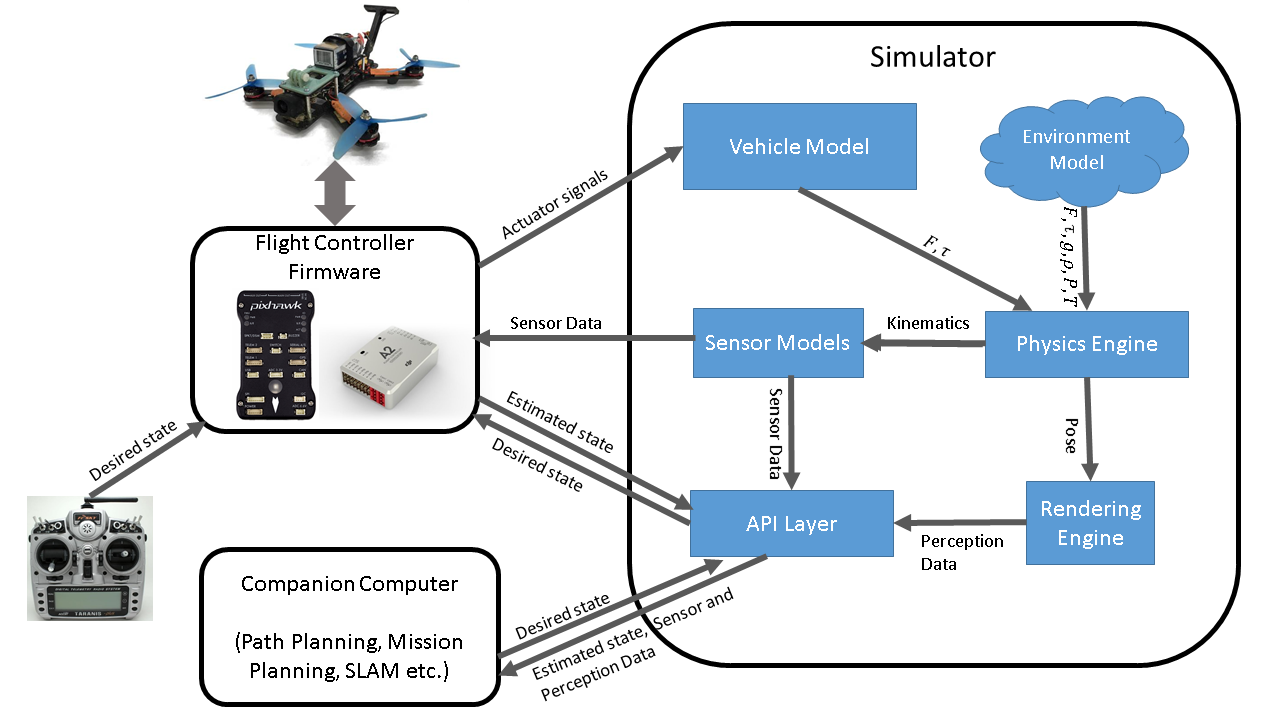}
		\caption{The architecture of the system that depicts the core components and their interactions.}
		\label{fig:arch}
	\end{figure}

	Our simulator follows a modular design with an emphasis on extensibility. The core components includes environment model, vehicle model, physics engine, sensor models, rendering interface, public API layer and an interface layer for vehicle firmware as depicted in Figure~\ref{fig:arch}.
	
	The typical setup for an autonomous aerial vehicle includes the flight controller firmware such as PX4 \cite{meier2011pixhawk}, ROSFlight \cite{jackson2016rosflight}, Hackflight\cite{levyHackflight} etc. The flight controller takes desired state and the sensor data as inputs, computes the estimate of current state and outputs the actuator control signals to achieve the desired state. For example, in case of quadrotors, user may specify desired pitch, roll and yaw angles as desired state and the flight controller may use sensor data from accelerometer and gyroscope to estimate the current angles and finally compute the motor signals to achieve the desired angles.
	
	During simulation, the simulator provides the sensor data from the simulated world to the flight controller. The flight controller outputs the actuator signals which is taken as input by the the vehicle model component of the simulator. The goal of the vehicle model is to compute the forces and torques generated by the simulated actuators. For example, in case of quadrotors, we compute the thrust and torques produced by the propellers given the motor voltages. In addition, there may be forces generated from drag, friction and gravity. These forces and torques are then taken as inputs by the physics engine to compute the next kinematic state of bodies in the simulated world. This kinematic state of bodies along with the environment models for gravity, air density, air pressure, magnetic field and geographic location (GPS coordinates) provides the ground truth for the simulated sensor models.
	
	The desired state input to the flight controller can be set by human operator using remote control or by a companion computer in the autonomous setting. The companion computer may perform expensive higher level computations such as determining next desired waypoint, performing simultaneous localization and mapping (SLAM), computing desired trajectory etc. The companion computer may have to process large amount of data generated by the sensors such as vision cameras and lidars which in turn requires that simulated environments have reasonable details. This has been one of the challenging areas where we leverage recent advances in rendering technologies implemented by platforms such as Unreal engine \cite{karis2013real}. In addition, we also utilize the underlying pipeline in the Unreal engine to detect collisions. The companion computer interacts with the simulator via a set of APIs that allows it to observe the sensor streams, vehicle state and send commands. These APIs are designed such that it shields the companion computer from being aware of whether its being run under simulation or in the real world. This is particularly important so that one can develop and test algorithms in simulator and deploy to real vehicle without having to make additional changes.
	
	The AirSim code base is implemented as a plugin for the Unreal engine that can be dropped in to any Unreal project. The Unreal engine platform offers an elaborate marketplace with hundreds of pre-made detailed environments, many created using photogrammetry techniques \cite{ue4openworld2015} to generate reasonably faithful reconstruction of real-world scenes.
	
	Next, we provide more details on the individual components of the simulator.
	
	\subsection{Vehicle Model}
	\label{sec:Vehicle}
	AirSim provides an interface to define vehicle as a rigid body that may have arbitrary number of actuators generating forces and torques. The vehicle model includes parameters such as mass, inertia, coefficients for linear and angular drag, coefficients of friction and restitution which is used by the physics engine to compute rigid body dynamics. 
	
	Formally, a vehicle is defined as a collection of $K$ vertices placed at positions $\{{\bf r}_1, .., {\bf r}_k\}$ and normals $\{{\bf n}_1, .., {\bf n}_k\}$, each of which experience a unitless vehicle specific scaler control input $\{u_1, .., u_k\}$. The forces and torques from these vertices are assumed to be generated in the direction of their normals. However note that the positions as well as normals are allowed to change during the simulation.
	
	\begin{figure}[ttbp]
		\includegraphics[width=0.93\textwidth]{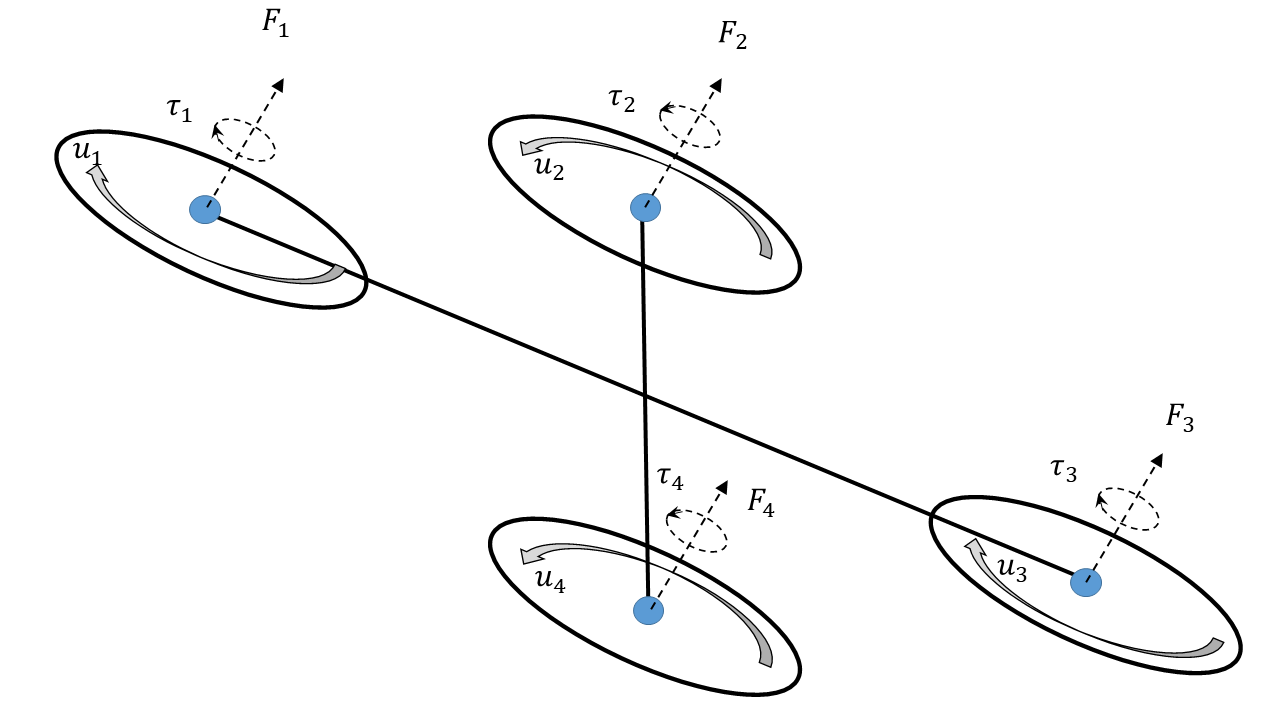}
		\caption{Vehicle model for the quadrotor. The four blue vertices experience the controls $u_1, .. u_4$, which in turn results in the forces ${\bf F}_1, .., {\bf F}_4$ and the torques ${\bf \tau}_1, .., {\bf \tau}_4$.}
		\label{fig:quad}
	\end{figure}
	Figure \ref{fig:quad} shows how a quadrotor can be depicted as a collection of four vertices. The control input $u_i$ drives the rotational speed of the propellers located at the four vertices. We compute the forces and torques produced by propellers using \cite{brandtuiuc}:
	\begin{align*}
	{\bf F}_i = C_T \rho \omega_{max}^2 D^4 u_i \mbox{\hspace{0.25in}and\hspace{0.25in}}
	{\bf \tau}_i = \frac{1}{2\pi}C_{pow} \rho \omega_{max}^2 D^5 u_i.
	\end{align*}
	Here $C_T$ and $C_{pow}$ are the thrust and the power coefficients respectively and are based on the physical characteristics of the propeller, $\rho$ is the air density, $D$ is the propeller's diameter and $\omega_{max}$ is the max angular velocity in revolutions per minute. By allowing the movements of these vertices during the flight it is possible to simulate the vehicles with capabilities such as Vertical Take-Off and Landing  (VTOL) and other recent quadrotors that change their configuration in flight.
	
	The vehicle model abstract interface also provides a way to specify the cross sectional area in body frame that in turn can be used by physics engine to compute the linear and angular drag on the body. 
	
	\subsection{Environment}
	The vehicle is exposed to various physical phenomena including gravity, air-density, air pressure and magnetic field. While it is possible to produce computationally expensive models of these phenomena that are very accurate, we focus our attention to models that are accurate enough to allow a real-time operation with hardware-in-the-loop. We describe these individual components of the environment below.
	
	\subsubsection{Gravity}
	While many models use a constant number to model the gravity, it varies in a complex manner as demonstrated by models such as GRACE \cite{tapley2007ggm03}. For most ground based or low altitude vehicles these variations may not be important; however, it is fairly inexpensive to incorporate a more accurate model. Formally, we approximate the gravitational acceleration $g$ at height $h$ by applying binomial theorem on Newton's law of gravity and neglecting the higher powers:
	\begin{equation*}
	g = g_{0}\cdot\frac{R_e^2}{(R_e + h)^2} \approx g_{0}\cdot \left (1 - 2\frac{h}{R_e} \right ).
	\end{equation*}
	Here $R_e$ is Earth's radius and $g_{0}$ is the gravitational constant measured at the surface.
	
	\subsubsection{Magnetic Field}
	\label{sec:env_magnetic}
	Accurately modeling the magnetic field of a complex body such as Earth is a computationally expensive task. The World Magnetic Model (WMM) model \cite{chulliat2015us} by National Oceanic and Atmospheric Administration (NOAA) is one of the best known magnetic models of Earth. Unfortunately, the most recent model WMM2015 is fairly complex and computationally expensive for real-time applications.
	
	We implemented the tilted dipole model where we assume Earth as a perfect dipole sphere \cite[pp 27-30]{lanza2006}. This ignores all but the first order terms to derive magnetic field estimate using the spherical geometry. This model allows us to simulate variation of the magnetic field as we move in space as well as areas that are often problematic such as polar regions. Given a geographic latitude $\theta$, longitude $\phi$ and altitude $h$ (from surface of the earth), we first compute the magnetic co-latitude $\theta_m$ using:
	\begin{equation*}
	\cos\theta_m = \cos\theta\cos\theta^0 + \sin\theta\sin\theta^0\cos(\phi-\phi^0).
	\end{equation*}
	Where $\theta^0$ and $\phi^0$ denote the latitude and longitude of the true magnetic north pole. Then, the total magnetic intensity $|B|$ is computed as:
	\begin{equation*}
	|B| = B_0 (\frac{R_e}{R_e + h})^3\sqrt{1 + 3\cos^2\theta_m}
	\end{equation*}
	Here $B_0$ is the mean value of the magnetic field at the magnetic equator on the Earth's surface, $\theta_m$ is the magnetic co-latitude and $R_e$ is the mean radius of the Earth. Next, we determine the inclination $\alpha$ and declination $\beta$ angles using: 
	\begin{equation*}
	\tan{\alpha} = 2\cot\theta_m \hspace{0.2in} \mbox{and} \hspace{0.2in} \sin\beta = 
	\begin{cases}
	\sin(\phi-\phi^0)\frac{\cos\theta^0}{\sin\theta_m}, & \mbox{if } \cos\theta_m > \sin\theta^0\sin\theta\\ 
	\cos(\phi-\phi^0)\frac{\cos\theta^0}{\sin\theta_m}, & \mbox{otherwise}.
	\end{cases}
	\end{equation*}
	Finally, we can compute the horizontal field intensity ($H$), the latitudinal ($X$), the longitudinal ($Y$) and the vertical field ($Z$) components of the magnetic field vector as follows:
	\begin{align*}
	H &= |B|\cos\alpha &\hspace{0.1in} Z &= |B|\sin\alpha &\hspace{0.1in} X &= H\cos\beta &\hspace{0.1in} Y &= H\sin\beta.
	\end{align*}
	
	\subsubsection{Air Pressure and Density}
	\label{sec:PressureAndAir}
	The relationship between the altitude and the pressure of the Earth's atmosphere is complicated due to the presence of many distinct layers, each with its own individual properties. First we compute Standard Temperature $T$ and Standard Pressure $P$ using 1976 U.S. Standard Atmosphere model \cite[eq 1.16, 1.17]{stull2016practical} for altitude below 51 kilometers and switch to the model in \cite[Table 4]{Braeunig} beyond that up to 86 km. Then, the air density is $\rho = \frac{P}{R \cdot T}$ (where $R$ is the specific gas constant.) 
	
	\subsection{Physics Engine}
	The kinematic state of the body is expressed using 6 quantities: position, orientation, linear velocity, linear acceleration, angular velocity and angular acceleration. The goal of the physics engine is to compute the next kinematic state for each body given the forces and torques acting on it. We strive for an efficient physics engine that can run its update loop at high frequency (1000 Hz) which is desirable for enabling real-time simulation scenarios such as high speed quadrotor control. Consequently, we implement a physics engine that avoids the extra complexities of a generic engine allowing us to tightly control the performance and make trade-offs that best meet our requirements.
	
	\subsubsection{Linear and Angular drag}
	Since the vehicle moves in the presence of air, the linear and the angular drag has a significant effect on the dynamics of the body. The simulator computes the magnitude $|{\bf F}_d|$ of the linear drag force on the body according to the drag equation \cite{taylor2005classical}:
	\begin{equation*}
	|{\bf F}_d| = \frac{1}{2}\rho |{\bf v}|^2 C_{lin} A.
	\end{equation*}
	Here $C_{lin}$ is the linear air drag coefficient, $A$ is the vehicle cross-section and $\rho$ is the air density. This drag force acts in the direction opposite to the velocity vector ${\bf v}$
	
	Computing the angular drag for arbitrary shape remains complex and computationally intensive task. Many existing physics engines use a small but often an arbitrary damping constant as a substitute for computing actual angular drag. We provide simple but better approximations to model the angular drag.
	
	Consider an infinitesimal surface area $ds$ in the extremity of the body experiencing the angular velocity ${\bf \omega}$. As the linear velocity ${\bf dv}$ experienced by $ds$ is given by ${\bf r}_{ds} \times {\bf \omega}$, we can now use the linear drag equation for $ds$ \cite[pp 160-161]{nakayama1998introduction}:
	\begin{equation*}
	|{\bf dF}| = \frac{1}{2}\rho |{\bf r}_{ds} \times {\bf \omega}|^{2} C_{lin} {ds}
	\mbox{,\hspace{0.25in} where direction of ${\bf dF}$ is $-{\bf r}_{ds} \times {\bf \omega}$}.
	\end{equation*}
	Now, the drag torque is computed by integrating over the entire surface: ${\bf \tau}_{d} = \int_{S} {\bf r}_{ds} \times {\bf dF}$.
	To simplify the implementation, we approximate the body of the vehicle as set of connected faces which further can be approximated as a rectangular box for the purpose of evaluating the integral.
	
	\subsubsection{Accelerations}
	In addition to the drag forces and torques, we also need to consider the forces ${\bf F}_i$ and the torques ${\bf \tau}_i$ present on the vehicle at the vertex located at ${\bf r}_i$  relative to center of gravity (see section~\ref{sec:Vehicle}). We thus compute the net force and torque as:
	
	\begin{align*}
	{\bf F}_{net} = \sum_{i} {\bf F}_{i} + {\bf F}_d\mbox{\hspace{0.25in}and\hspace{0.25in}} {\bf \tau}_{net} = \sum_{i} [{\bf \tau}_i + {\bf r}_{i} \times {\bf F}_{i}] + {\bf \tau}_d .
	\end{align*}
	We obtain the linear acceleration by applying Newton's second law and then adding gravity vector to compute the net acceleration,
	${\bf a} = {\bf F}_{net} / m + {\bf g}$. The angular acceleration in body frame is given by Euler's rotation equation:
	${\bf \alpha} = I^{-1} \cdot ({\bf \tau}_{net} - ({\bf \omega} \times (I \cdot {\bf \omega})))$,
	where, $I$ is the inertia tensor and ${\bf \omega}$ is angular velocity, both in body frame. 
	
	\subsubsection{Integration}
	We update the position ${\bf p}_{k+1}$ of the body at time $k+1$ by integrating the velocity and the initial position ${\bf p}_{0}$. The first order integration algorithms such as Euler method diverges quickly with unbounded error although very simple to implement. In our implementation we use Velocity Verlet algorithm instead of Runge Kutta for its computationally inexpensiveness and stability while still being second order method \cite{Herman2017}. Formally, 
	\begin{align*}
	{\bf v}_{k+1} &= {\bf v}_{k} + \frac{{\bf a}_{k} + {\bf a}_{k+1}}{2} \cdot dt &
	{\bf p}_{k+1} &= {\bf p}_{k} + {\bf v}_{k} \cdot dt + \frac{1}{2} \cdot {\bf a}_{k} \cdot dt^2
	\end{align*}
	The angular velocity is updated in similar manner as linear velocity however updating orientation isn't straight forward. One of the approach is to maintains the orientation as a rotation matrix that is updated every time step. However this causes a slow drift which must be corrected by orthonormalization at regular intervals which is expensive. Alternative approach is to maintain rotations as much more efficient quaternions which are also numerically stable and trivially normalizable. One of the problem, however, is that the orientation quaternion is maintained in the world frame while the angular velocity is maintained in the body frame in our framework. To update the orientation, we first compute the angle-axis pair $(\alpha_{dt}, {\bf u})$ where $\alpha_{dt}$ is the angle traversed around unit vector ${\bf u}$. We can compute the angle $\alpha_{dt} = |{\bf \omega}| \cdot dt$ and axis by $u = {\bf \omega} / |\omega|$. This allows us to compute equivalent change in quaternion ${\bf q}_{dt}$ representing the change in orientation in time $dt$. As noted before, ${\bf q}_{dt}$ is in body frame while ${\bf q}_{k}$ in world reference frame. The problem now remains that of adding ${\bf q}_{dt}$ to ${\bf q}_{k}$ to obtain ${\bf q}_{k+1}$ which can be proven to given by relationship ${\bf q}_{k+1} = {\bf q}_{k} \cdot {\bf q}_{dt}$.
	%
	%
	%
	%
	
	\subsubsection{Collisions}
	Unreal engine offers a rich collision detection system optimized for different classes of collision meshes and we directly use this feature for our needs. We receive the impact position, impact normal and penetration depth for each collision that occurred during the render interval. Our physics engine uses this data to compute the collision response with Coulomb friction to modify both linear and angular kinematics.\cite{hecker1997physics}
	
	\subsection{Sensors}
	AirSim offers sensor models for accelerometer, gyroscope, barometer, magnetometer and GPS. All our sensor models are implemented as C++ header-only library and can be independently used outside of AirSim. Like other components, sensor models are expressed as abstract interfaces so it is easy to replace or add new sensors.
	
	\subsubsection{Barometer}
	To simulate barometer, we compute ground truth pressure using the detailed model of atmosphere (sec~\ref{sec:PressureAndAir}) and model the drift in the pressure measurement over time using Gaussian Markov process \cite{sabatini2013stochastic} for more realistic behavior in long flights. Formally, if we denote the current bias factor as $b_k$ then the drift is modeled as:
	\begin{equation*}
	b_{k+1} = w\cdot b_k + (1-w)\cdot\eta, \mbox{where: } w = e^{- \frac{dt}{\tau}} \mbox{ and } \eta\sim N(0,s^2).
	\end{equation*}
	Here $\tau$, is the time constant for the process and set to $1$ hour in our model. $\eta$ is a zero mean Gaussian noise with standard deviation that can be selected using the data available in \cite{Burch2014}. This pressure $p$ is then added with white noise drawn from zero mean Gaussian distribution with standard deviation set from datasheet of the sensor (such as MEAS MS56112). Finally we convert the pressure to altitude using barometric formula used by the sensor's driver:
	\begin{equation*}
	h = \frac{T_{0}}{a} \left[\left(\frac{p}{p_{0}}\right)^{-(\frac{a \cdot R}{g})} - 1 \right],
	\end{equation*}
	here $T_{0}$ is the reference temperature (15 deg C), $a = -6.5 \times 10^{-3}$ is the temperature gradient, $g$ and $R$ are the gravity and the specific gas constants, $p_{0}$ is the current sea level pressure and $p$ is the measurement.
	
	\subsubsection{Gyroscope and Accelerometer}
	Gyroscope and accelerometers constitute the core of the inertial measurement unit (IMU) \cite{UCAM-CL-TR-696}. We model these by adding white noise and bias drift over time to the ground truth. For gyroscope, given the true angular velocity in body frame $\omega$, we compute the measurement $\omega^{\mbox{out}}$ as,
	\begin{align*}
	\omega^{\mbox{out}} &= \omega + \eta_a + b_t, &\mbox{where } &\eta_a\sim N(0, r_a) \mbox{ and } \\
	b_t &= b_{t-1} + \eta_b, &\mbox{where } &\eta_b\sim N \left(0, b_0\sqrt{\frac{dt}{t_a}} \right).
	\end{align*}
	Here parameters $r_a$, bias $b_0$ and the time constant for bias drift $t_a$ can either be obtained from Allan variance plots or from datasheets. Accelerometer output is computed in the similar manner except that we must first subtract gravity from the true linear acceleration in the world frame and then convert the result to the body frame before we add bias drift and noise.
	
	\subsubsection{Magnetometer}
	We use the tilted dipole model for Earth's magnetic field \ref{sec:env_magnetic}, given the geographic coordinates to compute the components of the ground truth magnetic field in body frame and add the white noise as specified in the datasheet.
	
	\subsubsection{Global Positioning System (GPS)}
	Our GPS model simulates latency (typically 200ms), slower update rates (typically 50 Hz) and horizontal and vertical position error estimate decay rates to simulate gaining fix over time. The decay rate is modeled using first order low pass filter individually parameterized for horizontal and vertical fix.
	
	\subsection{Visual Rendering}
	
	Since advanced rendering and detailed environments have been a key requirement for AirSim we chose Unreal Engine 4 (UE4) \cite{karis2013real} as our rendering platform. UE4 offers several features that made it an attractive choice including it being an open source and available on Linux, Windows as well as OSX. UE4 brings some of the cutting edge graphics features such as physically based materials, photometric lights, planar reflections, ray traced distance field shadows, lit translucency etc. Figure \ref{drone_depth_materials} shows a screen-shot from AirSim which highlight near photo-realistic rendering capabilities. Further, Unreal's large online Marketplace has various pre-made elaborate environments, many of which are created using photogrammetry techniques.

	\section{Experiments}
	We perform experiments primarily to evaluate how close the flight characteristic of a quadrotor flying in real-world is to that of a simulation of the same vehicle in AirSim. We also evaluate some of our sensor models against the real-world sensors.
	
	\vspace{0.075in}
	\noindent \textbf{Hardware Platform}: Real-world flights were performed with the Pixhawk v2 flight controller mounted on a Flamewheel quadrotor frame, together with a Gigabyte 5500 Brix running Ubuntu 16.04. The sensor measurements were recorded on the Pixhawk device itself. We configured the simulated quadrotor in AirSim using the measured physical parameters and simulated sensor models configured using sensor data sheets. The AirSim MavLinkTest application was used to perform repeatable offboard control for both the real-world and the simulated flights.
	
	\vspace{0.075in}
	\noindent \textbf{Trajectory Evaluation}: We fly the quadrotor in the simulator in two different patterns: (1) trajectory in square shape with each side being $5$m long (2) trajectory in circle shape with radius being $10$m long. We then use exact same commands to fly the real vehicle. For both the simulation and the real-world flights, we collect location of the vehicle in local NED coordinates along with timestamps.
	
\begin{figure}[t]
		\centering
		\subfigure[Circle maneuver]{\label{glamorshot_circle}\includegraphics[width=0.48\textwidth]{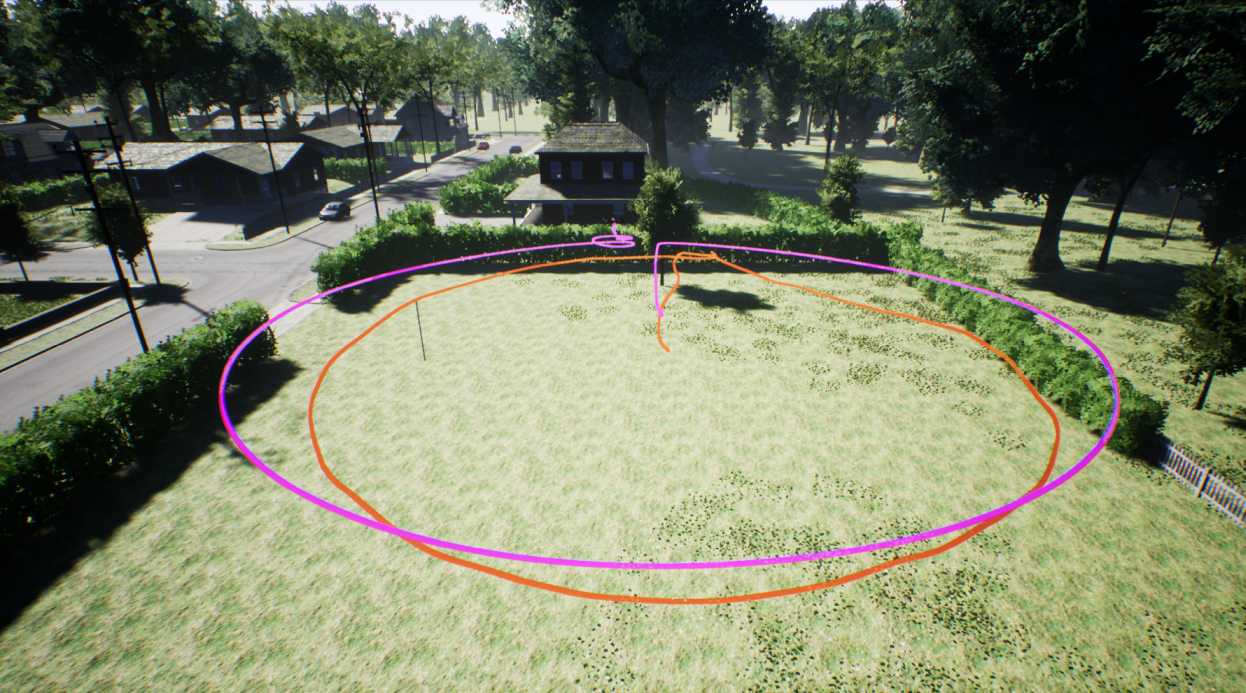}}
		\subfigure[Square maneuver]{\label{glamorshot_square}\includegraphics[width=0.48\textwidth]{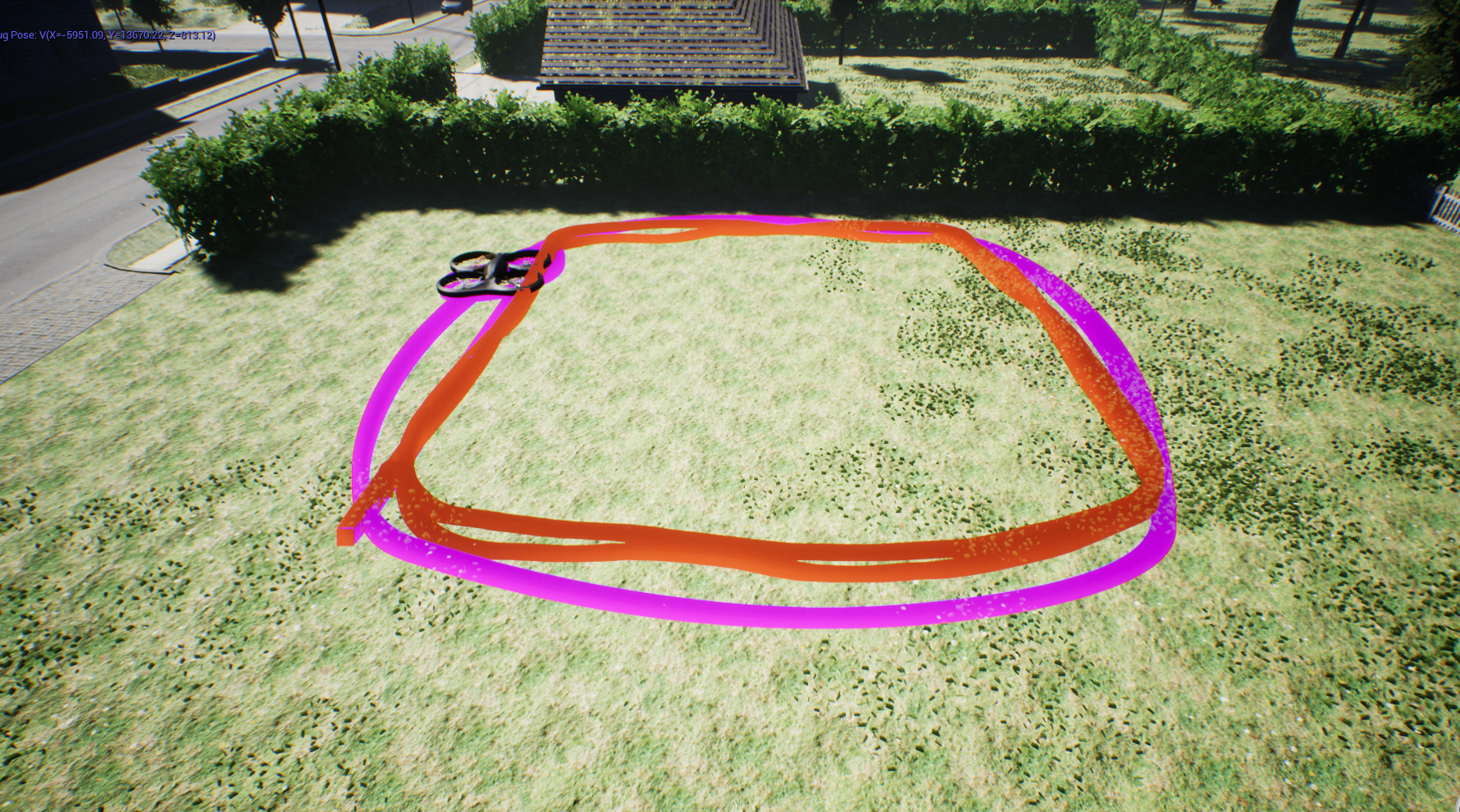}}
		\subfigure[Space-Time Plot for Circle]{\label{fig:circle_plot}\includegraphics[width=0.49\textwidth]{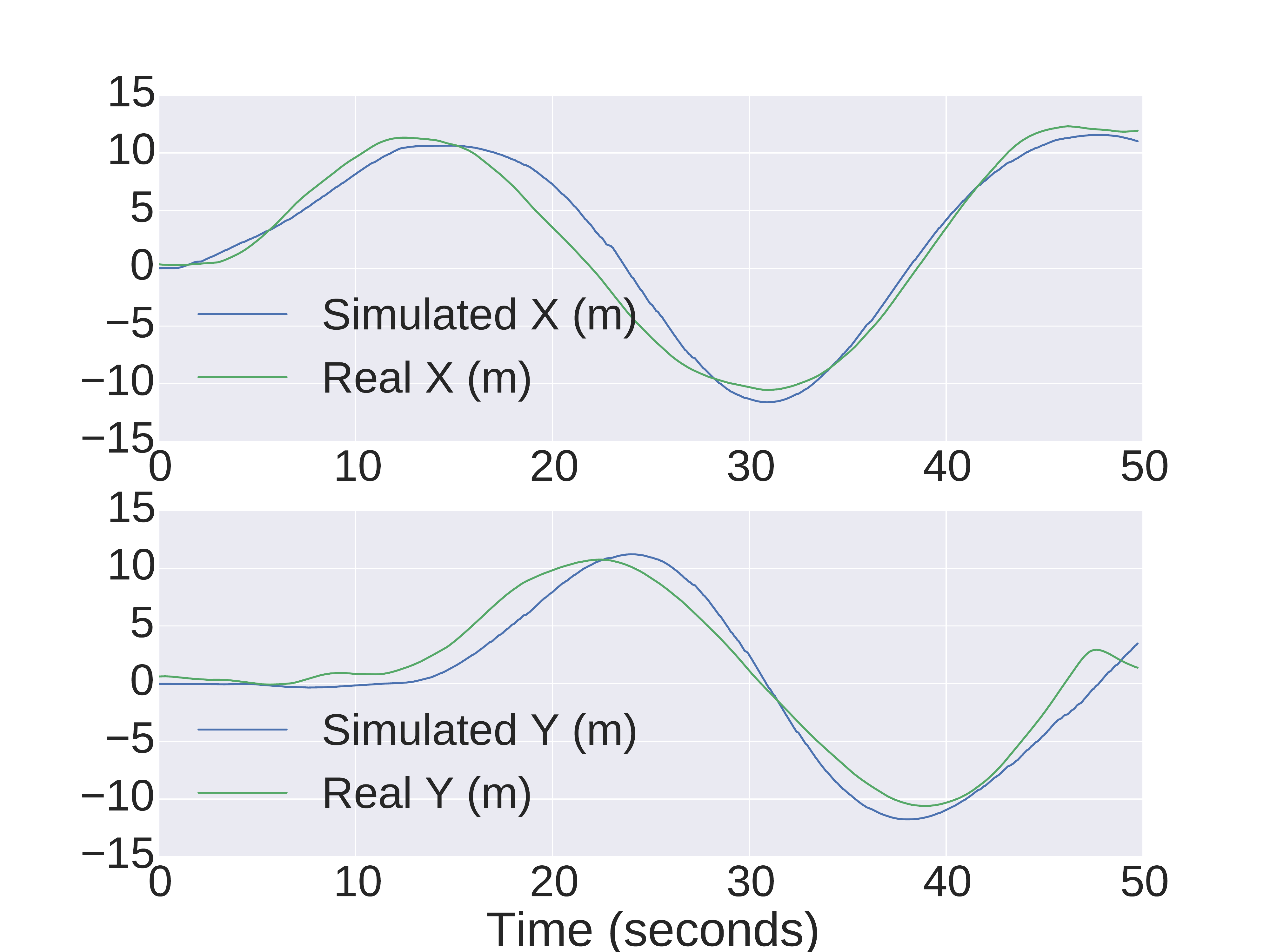}}
		\subfigure[Space-Time Plot for Square]{\label{fig:square_plot}\includegraphics[width=0.49\textwidth]{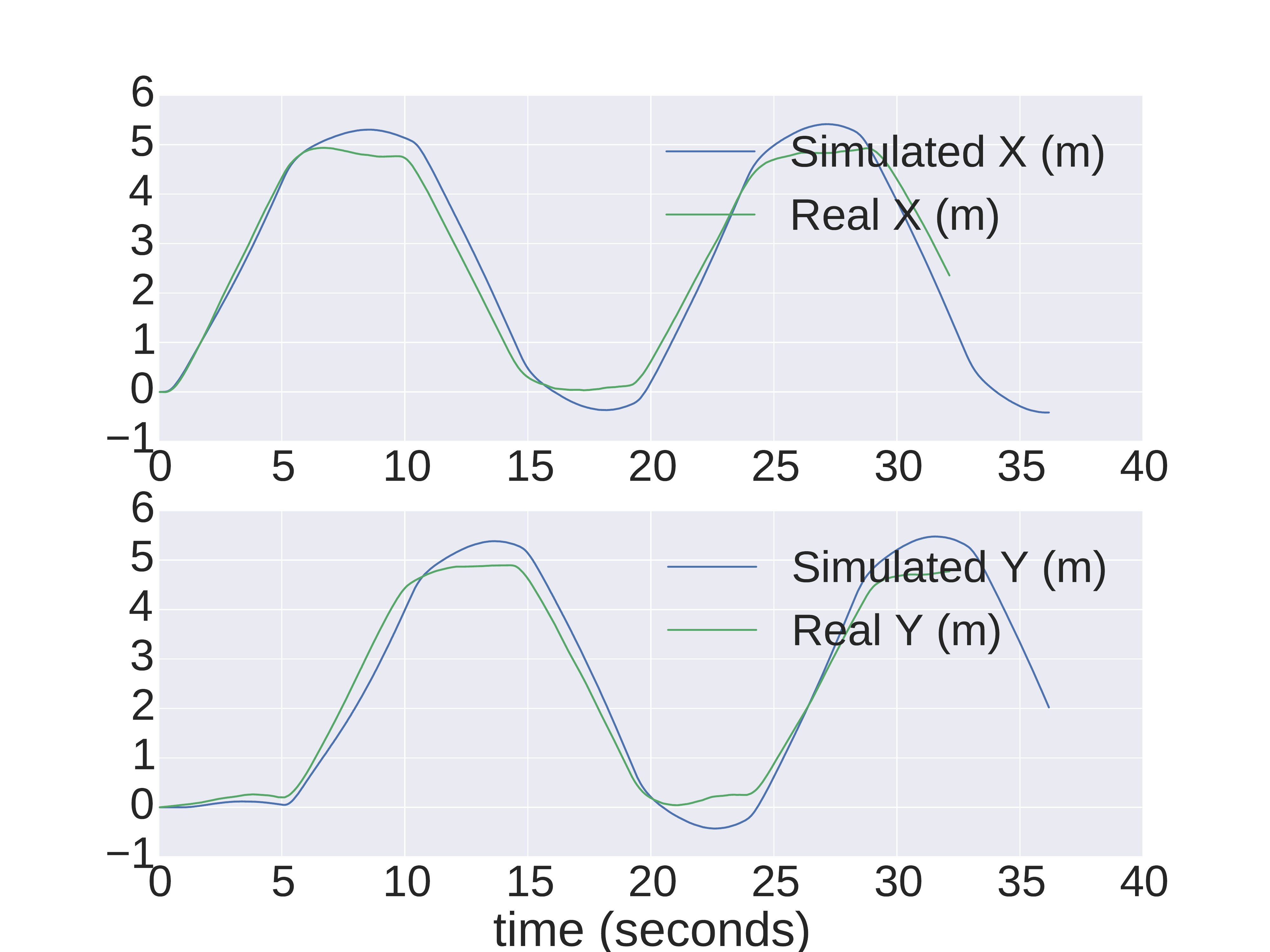}}
		\caption{Evaluating the differences between the simulated and the real-world flight. In top figures, the purple and the red lines depict the track from simulation and the real-world flights respectively.}
		\label{fig:traj_val}
	\end{figure}

	Figure \ref{fig:circle_plot} and \ref{fig:square_plot} shows the time series of locations in simulated flight and the real flight. Here, the horizontal axis represents the time and the vertical axis represent the off-set in X and Y directions. We also compute the symmetric Hausdorff distance between the real-world track and the track in simulation. We found that the simulation and real-world tracks were fairly close both for the circle (Hausdorff distance between simulated and real-world: $1.47$ m) as well as the square (Hausdorff distance between simulated and real-world: $0.65$ m).
	
	We also present visual comparison for this experiment for the circle and the square patterns in Figures \ref{glamorshot_circle} and \ref{glamorshot_square} respectively. The simulated trajectory is shown with a purple line while the real trajectory is shown with a red line. We can observe that qualitatively the trajectories tracked by both the real-world and the simulated vehicle are close. The small differences may have been caused by various factors such as integration errors, vehicle model approximations and mild random winds.
	
	\vspace{0.075in}
	\noindent \textbf{Sensor Models}: Besides evaluating the entire simulation pipeline we also investigated individual component models, namely the barometer (MEAS MS5611-01BA), the magnetometer (Honeywell HMC5883) and the IMU (InvenSense MPU 6000). Note that the simulated GPS model is currently simplistic, thus, we only focus on the three more complex sensor models. For each of the above sensors we use the manufacture specified datasheets to set the parameters in the sensor models. 
	
\begin{figure}[t]
		\centering
		\subfigure[Barometer]{\label{fig:baro_dyn}\includegraphics[width=0.48\textwidth]{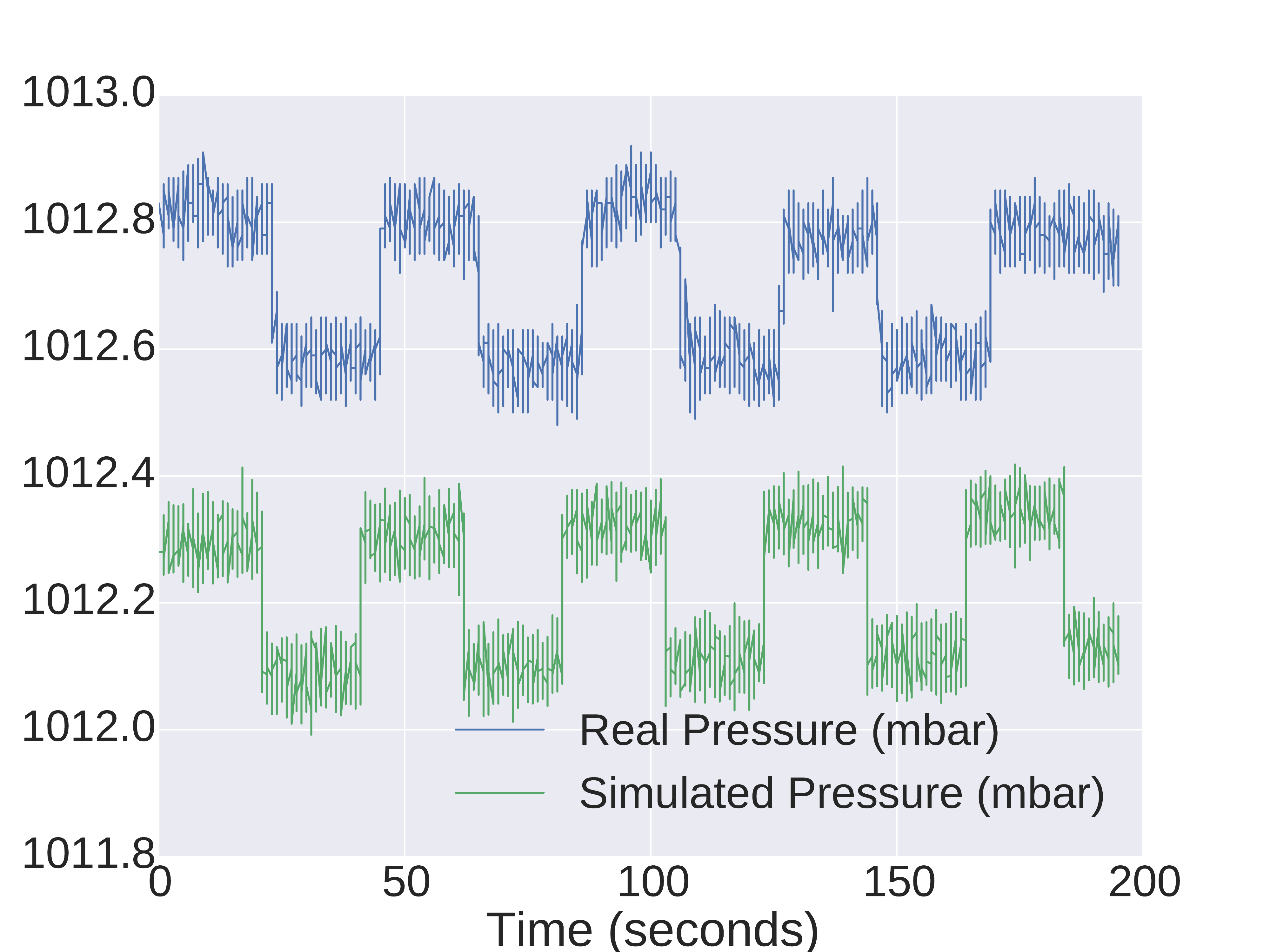}}
		\subfigure[Magnetometer]{\label{fig:mag_align}\includegraphics[width=0.48\textwidth]{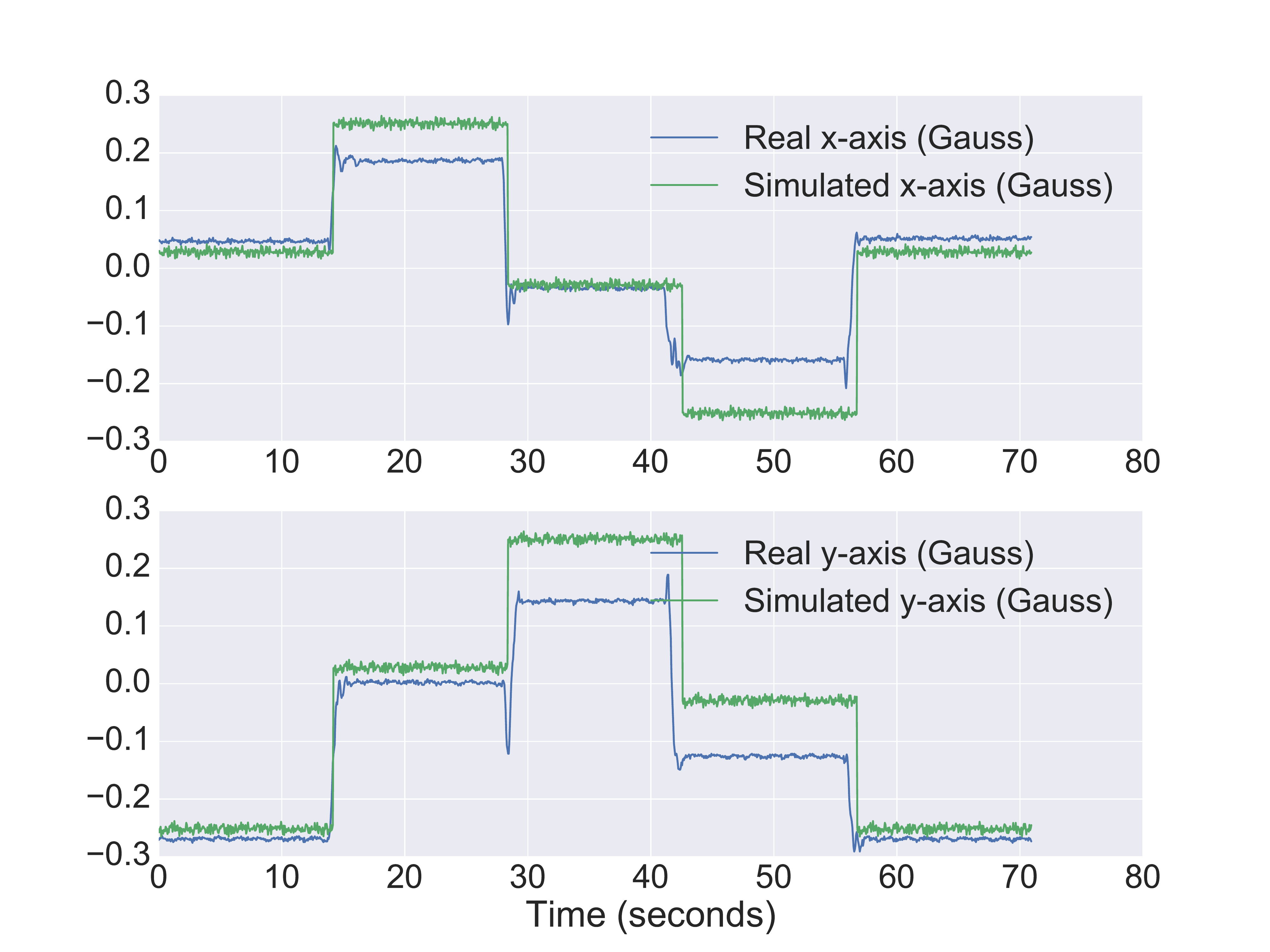}}
		\caption{Figure \ref{fig:baro_dyn} and \ref{fig:mag_align} show that barometer and the magnetometer characteristics in simulation closely resemble that of the real world.}
	\end{figure}

	\begin{itemize}
		\item {{\bf IMU:} We measured readings from the accelerometers and gyroscope as the vehicle was stationary and flying. We observed that while the characteristics were similar when the vehicle was stationary (gyro: simulated variance $\num{2.47E-07}$ $\mathrm{rad^2/s^2}$, real-world variance $\num{6.71E-07}$ $\mathrm{rad^2/s^2}$, accel.: simulated variance $\num{1.78E-04} $ $\mathrm{m^2/s^4}$, real-world variance $\num{1.93E-04}$ $\mathrm{m^2/s^4}$), the observed variance for an in-flight vehicle was much higher than the simulated one (accel.: simulated $\num{1.75E-3}$ $\mathrm{m^2/s^4}$ vs. real-world $\num{9.46}$ $\mathrm{m^2/s^4}$). This is likely in real-world the airframe vibrates when the motors are running and that phenomenon is not yet modeled in AirSim.}
		
		\item{{\bf Barometer:} We raised the sensor periodically between two fixed heights: ground level and then elevated to $178$ cm (both in simulation and real-world). Figure \ref{fig:baro_dyn} shows both the measurements (green is simulated, blue is real-world) and we observe that the signals have similar characteristics. Note that the offset between the simulated and the real-world pressure is due the difference in absolute pressure in the real-world and the one in the simulation. There is also a small increase in the middle due to a temperature increase, which wasn't simulated. Overall, the characteristics of the simulated sensor matches well to the real sensor.}
		
		\item{{\bf Magnetometer:} We placed the vehicle on the ground and then rotated it by $90^{\circ}$ four times. Figure \ref{fig:mag_align} shows the real-world and the simulated measurements and highlight that they are very similar in characteristic.}
	\end{itemize}
	
	\section{Conclusion and Future Work}
	AirSim offers hi-fidelity physical and visual simulation that allows to generate large quantity of training data cheaply for building machine learning models. AirSim API design allows developing algorithms against simulator and then deploy them without change on real vehicles. The core components of AirSim including physics engine, vehicle models, environment models and sensor models are designed to be independently usable with minimal dependencies outside of AirSim and are easily extensible. AirSim is inspired by the goal of developing reinforcement learning algorithms for the autonomous agents that can operate in the real world.
	
	The task of mimicking the real-world in \emph{real-time simulation} is a challenging endeavor. There are a number of things that can be improved. Currently we do not simulate richer collision response or advanced ground interaction models which may be possible in future by implementing our physics engine interface for NVIDIA PhysX and utilizing features such as physics sub-stepping. Also we do not simulate various oddities in camera sensors except those directly available in Unreal engine. We plan to add advanced noise models and lens models in future. The degradation of GPS signal due to obstacles is not simulated yet which we plan to add using ray tracing methods. We also plan to add more advanced wind effects and thermal simulations for fixed wing vehicles. Our extensibility APIs have been designed with above future work in mind and can also be used to realize other vehicle types.
	
	\bibliographystyle{spmpsci}
	{\footnotesize 
		\bibliography{references}}
	
\end{document}